\documentclass{article}

\usepackage{arxiv}

\usepackage[utf8]{inputenc} % allow utf-8 input
\usepackage[T1]{fontenc}    % use 8-bit T1 fonts
\usepackage{hyperref}       % hyperlinks
\usepackage{url}            % simple URL typesetting
\usepackage{booktabs}       % professional-quality tables
\usepackage{amsfonts}       % blackboard math symbols
\usepackage{amsmath}
\usepackage{bm}
\usepackage{nicefrac}       % compact symbols for 1/2, etc.
\usepackage{microtype}      % microtypography
\usepackage{graphicx}
\usepackage{float}
\usepackage[numbers,sort&compress]{natbib}
\usepackage{doi}
\usepackage{xcolor}
\usepackage{cleveref}

\newcommand{\phiMap}{\phi}
\DeclareMathOperator{\Diag}{Diag}
\newcommand{\githubmark}{\raisebox{-0.12em}{\includegraphics[height=0.95em]{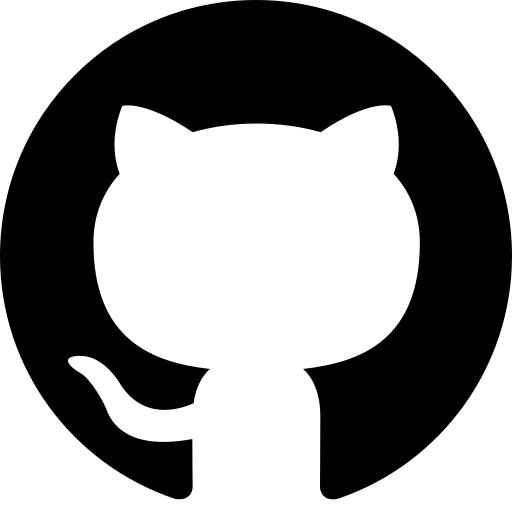}}}
\newcommand{\githubrepo}{\href{https://github.com/tommasocerruti/linear-attention-architectures}{\githubmark\hspace{1mm}tommasocerruti/linear-attention-architectures}}

\title{Linear Attention Architectures: Mechanisms, Trade-offs, and Cross-Layer Routing}

\author{ Tommaso Cerruti\textsuperscript{1,*} \\
	\And
	Tim Rieder\textsuperscript{1,*} \\
	\And
	George Rowlands\textsuperscript{1,*} \\
	\And
	Lingfeng Jin\textsuperscript{1,*} \\
	\And
	Imanol Schlag\textsuperscript{1,2} \\
	\\
	\parbox{\textwidth}{\centering \textsuperscript{1}ETH Zurich, D-INFK \quad \textsuperscript{2}ETH AI Center, ETH Zurich \\[0.5ex] \textsuperscript{*}Equal contribution} \\
}

\renewcommand{\headeright}{}
\renewcommand{\undertitle}{}
\renewcommand{\shorttitle}{Linear Attention Architectures}

\hypersetup{
pdftitle={Linear Attention Architectures: Mechanisms, Trade-offs, and Cross-Layer Routing},
pdfsubject={Comparison of efficient sequence-mixing architectures and cross-layer residual routing},
pdfauthor={Tommaso Cerruti, Tim Rieder, George Rowlands, Lingfeng Jin, Imanol Schlag},
pdfkeywords={Linear Attention, Recurrent Associative Memory, DeltaNet, Cross-layer Routing},
hidelinks,
}

\begin{document}
\maketitle

\begin{abstract}
Self-attention lets each token retrieve information from the full context, but its quadratic cost in sequence length limits training and inference at long context. This paper presents a comparative study of softmax attention and four recent recurrent linear-attention architectures: DeltaNet, Gated DeltaNet, Kimi Delta Attention, and Gated DeltaNet-2. We express these mechanisms in a common recurrent-memory notation, making explicit how they differ in expressivity, memory decay, erase and write control, training throughput, and implementation complexity. Our experiments center on 350M-parameter models trained for 15B tokens, and include optimizer and learning-rate comparisons, hybrid-versus-pure stack comparisons, sequence-length runtime measurements, larger DeltaNet runs at 1.3B and 3B parameters, and a small set of downstream evaluations. The reported speed results measure training throughput and iteration time; we do not provide an empirical inference-speed benchmark. Within the reported 350M-parameter, 15B-token sweep, Kimi Delta Attention with Muon reaches the lowest final validation loss, a pure Gated DeltaNet stack trained with AdamW has the highest normalized training throughput, hybrid stacks generally improve loss at a throughput cost, and Muon consistently lowers final validation loss relative to AdamW in the matched architecture settings we evaluate. We introduce and evaluate lightweight cross-layer routing mechanisms for DeltaNet-style memories. The most natural DeltaNet-inspired formulation, forwarding a lower layer's delta-rule write \emph{error} into the next layer's value target, does not improve over matched baselines. Routing into the aligned hidden stream and forwarding the write \emph{value} instead yields a modest improvement in the matched runs we report: \emph{Cross-Layer Value Routing (CLVR)} lowers final validation loss for both DeltaNet and Gated DeltaNet.
\end{abstract}

\begin{center}
	\githubrepo
\end{center}


\section{Introduction}

Transformer language models rely on self-attention~\cite{vaswani2017attention} to let each token retrieve information from other tokens in the context. This token-to-token information exchange, often called sequence mixing, is central to their expressivity. However, the same mechanism that makes softmax attention expressive also makes it expensive: computing the attention matrix requires explicit pairwise comparisons between every pair of tokens in the input sequence, leading to a cost that scales quadratically with sequence length. As models are deployed with larger context windows, this cost becomes a dominant factor in both training and inference, and motivates the search for sequence mixers with more favorable scaling.

\emph{Linear attention} is one of the most prominent responses to this challenge. By replacing the softmax kernel with a feature-map decomposition, linear attention can be reformulated as a recurrent update over a constant-size memory matrix, yielding linear-time training and constant-time inference per token. Early linear-attention variants traded a substantial amount of accuracy for this efficiency, but recent work has narrowed the gap considerably. In particular, DeltaNet~\cite{schlag2021linear} reinterprets linear attention as a fast-weight programmer and replaces the naive additive update with an error-correcting \emph{delta rule}: instead of writing the full value at every step, the model writes only the residual between the current value and what the fixed-size memory already predicts, reducing interference from overlapping key-value associations. Subsequent variants build on this idea by adding increasingly fine-grained mechanisms for controlling memory. Gated DeltaNet~\cite{yang2024gdn} introduces a learned scalar decay over the memory state, Kimi Delta Attention~\cite{kimi2025linear} refines this with a channel-wise decay gate, and Gated DeltaNet-2~\cite{hatamizadeh2026gdn2} further decouples the active delta-rule edit into separate channel-wise erase and write gates. Together, these architectures define a small but rapidly evolving family of recurrent memories that share a common skeleton but differ in how they balance selectivity, forgetting, and granularity of control.

Alongside these advances, a separate line of work has argued that deep language models suffer from \emph{information dilution}: as representations propagate through many layers, useful signals extracted at lower depths may become progressively harder to recover. Proposals such as Attention Residuals~\cite{attention_residuals} and Mixture-of-Depths Attention~\cite{mixture_of_depths_attention} address this by introducing explicit cross-layer pathways or depth-wise attention. These mechanisms are effective, but applying them directly to linear recurrent architectures partially defeats their efficiency advantages. This raises a natural question: is there a lightweight way to share information across depth that respects the linear-time structure of DeltaNet-style memories?

This paper makes four contributions:
\begin{enumerate}
    \item We express softmax attention, DeltaNet, Gated DeltaNet, Kimi Delta Attention, and Gated DeltaNet-2 in a common recurrent-memory notation. The framework isolates the role of the recurrent memory $W$, the delta-rule residual $r$, and the different decay, erase, and write mechanisms.

    \item We provide training-capable Megatron implementations and integrations of the newer linear-attention variants used in the study, including Kimi Delta Attention, Gated DeltaNet-2, and the cross-layer routing variants introduced here. The implementation is available at \githubrepo.

    \item We use this shared setup to map empirical trade-offs among the architectures, comparing validation loss, throughput, optimizer and learning-rate sensitivity, hybrid-versus-pure stack structure, sequence-length timing, larger DeltaNet runs, and downstream behavior.

    \item We introduce lightweight cross-layer routing for DeltaNet-style memories. Starting from \emph{Cross-Layer Error Residuals (CLER)}, which forward a lower layer's delta-rule write error into the next layer's value target, we find that this DeltaNet-inspired formulation does not improve over matched baselines. We then route into the aligned hidden stream and find that the layer's write \emph{value}, rather than its write error, is the useful signal. The resulting method, \emph{Cross-Layer Value Routing (CLVR)}, gives a small reduction in final validation loss in the matched DeltaNet and Gated DeltaNet runs we report, while preserving the host architecture's linear-time structure.
\end{enumerate}

We organize the comparison along three axes:
\begin{itemize}
    \item \emph{Mechanism}: we compare the algebraic structure of each recurrent update.
    \item \emph{Trade-offs}: we discuss how each design balances selectivity, memory decay, erase and write control, and memory granularity.
    \item \emph{Empirical behavior}: we report validation-loss, throughput, sequence-length timing, and downstream results across the variants we study.
\end{itemize}

Our aim is not to declare a single best architecture, but to make the design space legible. Viewing these mechanisms side by side makes it easier to see which choices are responsible for which properties, where scaling behavior is strongest, and which cross-layer routing variants remain open questions for future evaluation.

\section{Background: From Softmax Attention to Linear Attention}

This section reviews the path from standard softmax attention to linear attention, and introduces the recurrent-memory perspective that underlies all of the DeltaNet-style architectures studied in this paper. We focus on the algebraic structure rather than implementation details, and use the same single-head notation that will be used throughout the rest of the paper.

\paragraph{Softmax attention.}
Given queries and keys $q^{(i)}, k^{(i)} \in \mathbb{R}^{d_k}$ and values $v^{(i)} \in \mathbb{R}^{d_v}$ for $i = 1, \dots, T$, causal softmax attention computes the output at position $i$ as
\begin{equation}
\label{eq:softmax-attention}
y^{(i)}
=
\sum_{j \leq i}
\frac{
\exp\!\left( q^{(i)\top} k^{(j)} / \sqrt{d_k} \right)
}{
\sum_{\ell \leq i}
\exp\!\left( q^{(i)\top} k^{(\ell)} / \sqrt{d_k} \right)
}
\, v^{(j)}.
\end{equation}
The current query is explicitly compared against every previous key, and the values are combined via a normalized weighted average. This formulation is highly expressive: each token can in principle retrieve information from any previous position, and the softmax normalization gives the mechanism a built-in form of soft selection. However, the same explicit comparison is also the source of its quadratic cost. Training on a sequence of length $T$ requires computing the $\mathcal{O}(T^2)$ query-key interaction matrix, and storing the full attention matrix also requires $\mathcal{O}(T^2)$ memory. For long contexts, both the time and the memory cost become limiting.

\paragraph{The kernel view.}
Linear attention arises from a simple observation: if the unnormalized attention weight could be written as an inner product between feature maps of the query and the key, then the sum over previous positions could be rearranged to avoid the explicit pairwise comparison. Concretely, suppose there is a feature map $\phiMap : \mathbb{R}^{d_k} \to \mathbb{R}^{d_\phi}$ such that the (unnormalized) similarity can be approximated as
\begin{equation}
\exp\!\left( q^{(i)\top} k^{(j)} / \sqrt{d_k} \right)
\;\approx\;
\phiMap\!\left(q^{(i)}\right)^{\top} \phiMap\!\left(k^{(j)}\right).
\end{equation}
Substituting this into the unnormalized numerator of the attention output and exchanging the order of summation gives
\begin{equation}
\sum_{j \leq i}
\phiMap\!\left(q^{(i)}\right)^{\top} \phiMap\!\left(k^{(j)}\right) v^{(j)}
=
\phiMap\!\left(q^{(i)}\right)^{\top}
\underbrace{\sum_{j \leq i} v^{(j)} \otimes \phiMap\!\left(k^{(j)}\right)}_{\text{accumulated memory}}.
\end{equation}
The bracketed sum no longer depends on the query and can be maintained incrementally as $i$ advances. The query then interacts with a single matrix rather than with all previous tokens individually.

\paragraph{Recurrent memory.}
Defining the running sum as a memory matrix
\begin{equation}
W^{(i)}
\;=\;
\sum_{j \leq i} v^{(j)} \otimes \phiMap\!\left(k^{(j)}\right)
\;\in\; \mathbb{R}^{d_v \times d_\phi},
\end{equation}
the (unnormalized) output of linear attention can be written as a recurrence:
\begin{equation}
W^{(i)} = W^{(i-1)} + v^{(i)} \otimes \phiMap\!\left(k^{(i)}\right),
\qquad
y^{(i)} = W^{(i)} \, \phiMap\!\left(q^{(i)}\right).
\end{equation}
Each token contributes an additive write to the memory, and each output is a single matrix-vector product. The cost per token is constant in the sequence length, and total training cost is linear in $T$. The memory matrix can be interpreted in two equivalent ways: as a compressed representation of all previous (key, value) pairs, or, following the fast-weight programmer view~\cite{schlag2021linear}, as a set of \emph{fast weights} that are written to and read by the surrounding network.

\paragraph{What linear attention gives up.}
The efficiency gains come at a price. Softmax attention performs a per-query normalization that effectively re-weights the contributions of all previous tokens, and its exponential kernel makes the attention distribution sharp. Linear attention, in its naive additive form, has neither property. The memory $W^{(i)}$ accumulates previous writes without a learned decay or erase mechanism, so old associations remain active unless they are overwritten indirectly. There is also no built-in mechanism for sharpening the retrieval distribution or for forgetting stale information. As more (key, value) pairs are written into a fixed-size memory, interference between stored associations grows: a query that should retrieve one value may also pick up spurious contributions from unrelated keys that happen to be correlated with it under $\phiMap$. This interference is the central failure mode that more recent linear-attention variants are designed to address.

\paragraph{From additive writes to delta-rule writes.}
DeltaNet and its successors can be understood as principled answers to the interference problem. Instead of always adding the raw value $v^{(i)}$ to memory, these architectures first ask what the current memory already predicts for the current key,
\begin{equation}
\bar v^{(i)} \;=\; W^{(i-1)} \phiMap\!\left(k^{(i)}\right),
\end{equation}
and then write only the residual $r^{(i)} = v^{(i)} - \bar v^{(i)}$. This converts the memory update from a pure accumulator into an error-correcting writer, and gives the architecture a notion of \emph{what the memory does not yet know}. Subsequent variants augment this delta-rule update with mechanisms for forgetting and editing, ranging from a single scalar decay gate to channel-wise decay and finally to separate channel-wise erase and write gates. These additions give the model explicit control over how aggressively old information is decayed, which associations are erased, and which value channels are committed to memory. The next section formalizes these architectures in a common notation and makes their differences precise.

\section{Architectures}

We describe all attention mechanisms in a single-head notation and omit output projections, normalization layers, and feed-forward blocks for clarity. At token position $i$, the input representation $x^{(i)}$ is mapped to a query, key, and value vector,
\[
q^{(i)}, k^{(i)} \in \mathbb{R}^{d_k},
\qquad
v^{(i)} \in \mathbb{R}^{d_v}.
\]
For the linear-attention variants, keys and queries are passed through a feature map $\phiMap(\cdot)$, and the recurrent state is represented by a matrix
\[
W^{(i)} \in \mathbb{R}^{d_v \times d_\phi}.
\]
We interpret $W^{(i)}$ as the main associative memory storing key-value information up to token $i$. In this orientation, the memory maps transformed keys to values, so a read at key $k^{(i)}$ returns a value-space vector. For compactness, we write
\begin{equation}
\label{eq:feature-mapped-key}
\kappa^{(i)}
=
\phiMap\!\left(k^{(i)}\right)
\in \mathbb{R}^{d_\phi}
\end{equation}
for the feature-mapped key.

After a recurrent-memory variant has updated $W^{(i)}$, its token output is obtained by querying the memory with the transformed query:
\begin{equation}
\label{eq:common-output-read}
y^{(i)}
=
W^{(i)}
\phiMap\!\left(q^{(i)}\right).
\end{equation}

Several quantities are shared across the DeltaNet-style variants. Using the feature-mapped key from Eq.~\eqref{eq:feature-mapped-key}, we define the memory prediction at the current key as
\begin{equation}
\label{eq:memory-prediction}
\bar v^{(i)}
=
W^{(i-1)} \kappa^{(i)},
\end{equation}
and the corresponding delta-rule residual as
\begin{equation}
\label{eq:delta-residual}
r^{(i)}
=
v^{(i)} - \bar v^{(i)}.
\end{equation}
Thus, $r^{(i)}$ measures the part of the current value that is not already predicted by the existing memory.

The scalar gates used by the recurrent variants are
\begin{equation}
\label{eq:scalar-gates}
\alpha^{(i)}
=
f_{\alpha}\!\left(x^{(i)}\right)
\in (0,1),
\qquad
\beta^{(i)}
=
\sigma\!\left(w_{\beta}^{\top} x^{(i)}\right)
\in (0,1).
\end{equation}
Here, $\alpha^{(i)}$ is a learned token-dependent decay factor, while $\beta^{(i)}$ controls the write strength of the main memory.

For Kimi Delta Attention and Gated DeltaNet-2, we additionally use a vector-valued forget gate
\begin{equation}
\label{eq:channel-forget-gate}
\bm{\alpha}^{(i)}
=
f_{\bm{\alpha}}\!\left(x^{(i)}\right)
\in (0,1)^{d_\phi},
\qquad
D_{\alpha}^{(i)}
=
\Diag\!\left(\bm{\alpha}^{(i)}\right)
\in
\mathbb{R}^{d_\phi \times d_\phi}.
\end{equation}
This can be viewed as a channel-wise analogue of the scalar decay gate, applied along the transformed-key dimension of the recurrent memory.

For Gated DeltaNet-2, we also use a channel-wise erase gate and a channel-wise write gate,
\begin{equation}
\label{eq:gdn2-erase-write-gates}
\bm b^{(i)}
=
\sigma\!\left(W_b x^{(i)}\right)
\in (0,1)^{d_\phi},
\qquad
\bm w^{(i)}
=
\sigma\!\left(W_w x^{(i)}\right)
\in (0,1)^{d_v}.
\end{equation}
The erase gate controls which transformed-key channels are used to remove old content from memory, while the write gate controls which value channels are committed to memory.

\subsection{Softmax Attention}

Standard causal softmax attention, defined in Eq.~\eqref{eq:softmax-attention}, directly compares the current query with all previous keys and forms a normalized weighted average of the corresponding values.

Softmax attention is highly expressive because each token can selectively retrieve information from the entire previous context. However, this explicit pairwise comparison leads to quadratic cost in the sequence length during training. Linear-attention variants replace the explicit attention matrix with recurrent memory states that can be updated incrementally.

\subsection{DeltaNet}

DeltaNet replaces naive additive linear attention with an error-correcting delta rule~\cite{schlag2021linear}. Using the memory prediction $\bar v^{(i)}$ and residual $r^{(i)}$ from Eqs.~\eqref{eq:memory-prediction} and~\eqref{eq:delta-residual}, the memory is updated by writing only this residual:
\begin{equation}
W^{(i)}
=
W^{(i-1)}
+
\beta^{(i)}
r^{(i)}
\otimes
\kappa^{(i)},
\end{equation}
and the output is computed using the shared read rule in Eq.~\eqref{eq:common-output-read}.

The key idea is that the model does not simply add the new value to memory. Instead, it first asks what the memory already predicts for the current key, and then writes the correction needed to move the stored association toward $v^{(i)}$. This makes the update selective and key-specific.

The main advantage of DeltaNet is that it improves over naive additive storage by using an error-correcting update. Its limitation is that it has no explicit mechanism for globally clearing stale information. As interference accumulates, the model can correct individual associations, but it cannot directly decay the previous memory state in a coarse way.

\subsection{Gated DeltaNet}

Gated DeltaNet augments DeltaNet with a learned scalar forgetting mechanism~\cite{yang2024gdn}. Before computing the delta-rule residual, the previous memory is decayed by the token-dependent scalar gate $\alpha^{(i)}$ from Eq.~\eqref{eq:scalar-gates}. The decay-adjusted prediction is
\begin{equation}
\bar v_{\alpha}^{(i)}
=
\alpha^{(i)}
W^{(i-1)}
\kappa^{(i)}
=
\alpha^{(i)}
\bar v^{(i)},
\end{equation}
and the corresponding residual is
\begin{equation}
r_{\alpha}^{(i)}
=
v^{(i)}
-
\bar v_{\alpha}^{(i)}.
\end{equation}
The memory update is then
\begin{equation}
W^{(i)}
=
\alpha^{(i)} W^{(i-1)}
+
\beta^{(i)}
r_{\alpha}^{(i)}
\otimes
\kappa^{(i)},
\end{equation}
with output again computed by Eq.~\eqref{eq:common-output-read}.

Gated DeltaNet preserves the delta-rule correction, but applies it relative to a decayed version of the previous memory. This gives the architecture an explicit way to forget information, which can help reduce interference in long or cluttered contexts.

The trade-off is that the forgetting operation is state-wide. DeltaNet is purely selective in the sense that it updates memory through a key-specific residual. Gated DeltaNet keeps this corrective write, but also introduces the coarse decay term $\alpha^{(i)} W^{(i-1)}$. As a result, it gains a direct mechanism for memory clearance, but partially sacrifices the strictly selective character of the original DeltaNet update.

\subsection{Kimi Delta Attention}

Kimi Delta Attention keeps the gated delta-rule structure but replaces scalar forgetting with channel-wise forgetting~\cite{kimi2025linear}. Instead of applying the same decay factor to the entire memory, the model uses the vector-valued gate $\bm{\alpha}^{(i)}$ from Eq.~\eqref{eq:channel-forget-gate} to decay different transformed-key dimensions at different rates.

The channel-wise decay-adjusted prediction is
\begin{equation}
\bar v_{\bm{\alpha}}^{(i)}
=
W^{(i-1)}
D_{\alpha}^{(i)}
\kappa^{(i)},
\end{equation}
with residual
\begin{equation}
r_{\bm{\alpha}}^{(i)}
=
v^{(i)}
-
\bar v_{\bm{\alpha}}^{(i)}.
\end{equation}
The update becomes
\begin{equation}
W^{(i)}
=
W^{(i-1)}
D_{\alpha}^{(i)}
+
\beta^{(i)}
r_{\bm{\alpha}}^{(i)}
\otimes
\kappa^{(i)},
\end{equation}
and the output is computed by Eq.~\eqref{eq:common-output-read}.

Because $W$ maps transformed-key features to values, the diagonal decay matrix multiplies on the right of $W$ and acts along the key-feature dimension. This mechanism can be viewed as a more fine-grained version of Gated DeltaNet. Rather than decaying the whole memory with a single scalar, Kimi Delta Attention can preserve some feature dimensions while forgetting others more aggressively. If the vector gate collapses to a scalar gate,
\[
\bm{\alpha}^{(i)}
=
\alpha^{(i)} \mathbf{1}_{d_\phi},
\]
then the update reduces to the scalar-forgetting form used by Gated DeltaNet.

The advantage of this formulation is that it increases the granularity of memory control. Relative to Gated DeltaNet, it can distinguish between feature subspaces that should be retained and feature subspaces that should be cleared. However, the active delta-rule edit is still controlled by a single scalar $\beta^{(i)}$. The same gate controls how much old content is removed at the current key and how much new value is written into memory.

\subsection{Gated DeltaNet-2}

Gated DeltaNet-2 extends Kimi Delta Attention by decoupling the scalar delta gate into the channel-wise erase gate $\bm b^{(i)}$ and channel-wise write gate $\bm w^{(i)}$ from Eq.~\eqref{eq:gdn2-erase-write-gates}~\cite{hatamizadeh2026gdn2}. Instead of using the same scalar $\beta^{(i)}$ to control both removal of old content and insertion of new content, $\bm b^{(i)}$ controls the key-side erase operation, while $\bm w^{(i)}$ controls the value-side write operation.

Using the channel-wise decay matrix from Eq.~\eqref{eq:channel-forget-gate}, the decay-adjusted memory is
\begin{equation}
\widetilde W^{(i-1)}
=
W^{(i-1)}
D_{\alpha}^{(i)}.
\end{equation}
Gated DeltaNet-2 then defines a gated erase direction and a gated write target:
\begin{equation}
e^{(i)}
=
\bm b^{(i)} \odot \kappa^{(i)},
\qquad
z^{(i)}
=
\bm w^{(i)} \odot v^{(i)}.
\end{equation}
The residual written to memory is
\begin{equation}
r_{\mathrm{GDN2}}^{(i)}
=
z^{(i)}
-
\widetilde W^{(i-1)} e^{(i)}.
\end{equation}
The memory update is
\begin{equation}
W^{(i)}
=
\widetilde W^{(i-1)}
+
r_{\mathrm{GDN2}}^{(i)}
\otimes
\kappa^{(i)},
\end{equation}
and the output is computed by Eq.~\eqref{eq:common-output-read}.

Thus, Gated DeltaNet-2 preserves the channel-wise decay of Kimi Delta Attention, but makes the active delta update more flexible. The erase gate decides which key-feature channels should be removed from the previous memory, while the write gate decides which value channels should be stored. If
\[
\bm b^{(i)}
=
\beta^{(i)}\mathbf{1}_{d_\phi},
\qquad
\bm w^{(i)}
=
\beta^{(i)}\mathbf{1}_{d_v},
\]
then the update reduces to Kimi Delta Attention. If the decay gate also collapses to
\[
\bm{\alpha}^{(i)}
=
\alpha^{(i)}\mathbf{1}_{d_\phi},
\]
then it further reduces to Gated DeltaNet.

\section{Cross-Layer Routing}
\label{sec:cler}

Deep stacks can dilute information extracted at lower layers, and cross-layer schemes such as Attention Residuals and Mixture-of-Depths Attention~\cite{attention_residuals,mixture_of_depths_attention} address this with explicit depth-wise pathways. We pursue the same goal for DeltaNet-style associative memories, but without a new depth-attention operator: we ask whether an internal signal already produced inside the delta-rule update can be reused across depth as a lightweight, linear-time pathway. We explore two axes: \emph{what} to route (the layer's write error or its write value) and \emph{where} to inject it (the next layer's value target or the shared residual stream). We use three names throughout: \emph{Cross-Layer Error Residuals (CLER)} is the initial value-target error-routing formulation; \emph{CLER-H} routes the same write error into the shared hidden stream; and \emph{Cross-Layer Value Routing (CLVR)} routes the write value into the shared hidden stream. This keeps the ablation order explicit: CLER changes neither signal nor space, CLER-H fixes the injection space, and CLVR keeps the hidden-stream injection while changing the routed signal.

All routing variants share the same footprint. They augment existing DeltaNet variants with a side-channel that carries an internal signal from one recurrent-memory layer to a higher one, and they leave the underlying mixer, the recurrent update rule, and the gating unchanged. They differ only in which internal signal is carried and where it is added, which lets matched experiments isolate the effect of each choice rather than confounding it with a change to the host architecture.

In this section, a \emph{routing-capable layer} means a DeltaNet-style recurrent-memory layer that both produces a delta-rule write residual and can receive a routed residual from a lower such layer. Softmax-attention layers are not routing-capable in the reported hybrid stacks: they may pass the side-channel forward, but they do not produce or consume routed signals. The notation in this section switches from the generic token index $i$ to $(l,t)$ in order to track both layer depth $l$ and sequence position $t$.

\subsection{Delta-rule write quantities}

In all cross-layer routing variants we implement, each DeltaNet or Gated DeltaNet layer maintains a recurrent associative memory $W_{l,t}\in\mathbb{R}^{d_v\times d_\phi}$ that predicts the current value from the current key. For a layer $l$ and sequence position $t$, the model first computes a feature-mapped key and a prediction from the previous memory state:
\begin{equation}
\kappa_{l,t}=\phiMap(k_{l,t})\in\mathbb{R}^{d_\phi},
\qquad
\bar{v}_{l,t} = W_{l,t-1}\kappa_{l,t}\in\mathbb{R}^{d_v},
\end{equation}
where $\phiMap(\cdot)$ denotes the feature map used by the linear attention rule. The layer then computes a write residual:
\begin{equation}
r_{l,t} = v_{l,t} - \bar{v}_{l,t}\in\mathbb{R}^{d_v},
\end{equation}
which represents the correction that must be written into memory. These two quantities, the write residual $r_{l,t}$ (the part of the value the memory has not yet absorbed) and the write value $v_{l,t}$ itself, are the internal signals carried across depth: CLER and CLER-H use the residual, while CLVR uses the value.

\subsection{Cross-Layer Error Residuals (CLER)}

\begin{figure}[tbp]
    \centering
    \includegraphics[width=0.85\linewidth]{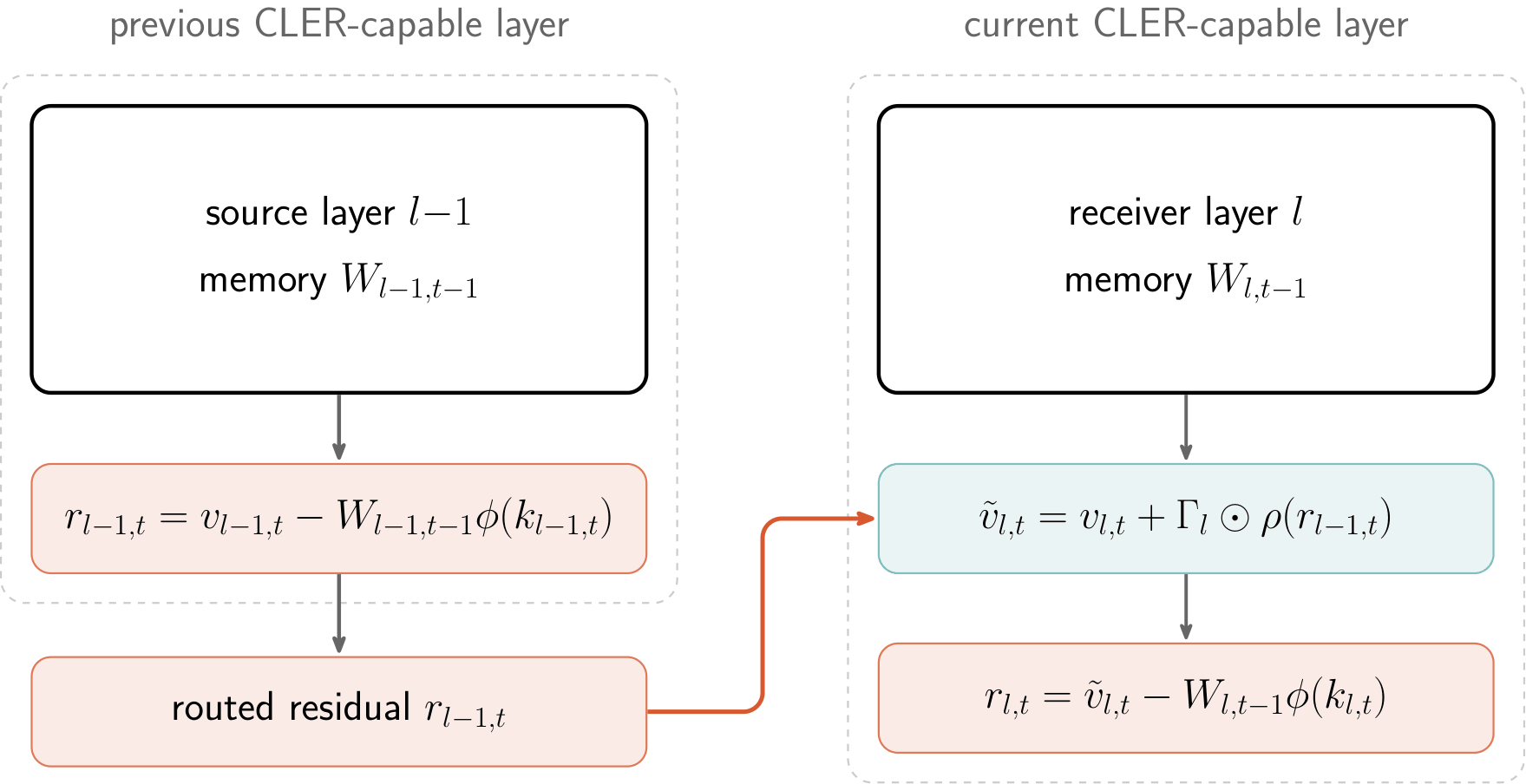}
    \caption{Cross-Layer Error Residuals (CLER). A lower routing-capable recurrent-memory layer computes a DeltaNet-style write residual. If softmax-attention layers intervene, the residual is carried forward as a side-channel. The next routing-capable layer scales the routed residual and injects it into its value target before computing its own write residual.}
    \label{fig:architecture-diagram}
\end{figure}

The direct way to share this correction across depth, inspired by DeltaNet-style architectures, is to forward the write error itself. This is the \emph{Cross-Layer Error Residuals (CLER)} formulation (\Cref{fig:architecture-diagram}): it injects the residual from the nearest lower routing-capable layer into the current layer's value target before the new residual is computed:
\begin{equation}
\label{eq:cler-value-target}
\tilde{v}_{l,t}
=
v_{l,t}
+
\Gamma_l \rho(r_{p(l),t}),
\end{equation}
where $p(l)$ denotes the nearest lower DeltaNet/Gated DeltaNet layer that produced a CLER residual, $\Gamma_l\in\mathbb{R}$ is a learned per-layer scalar, and $\rho:\mathbb{R}^{d_v}\to\mathbb{R}^{d_v}$ is a residual normalization taken as the identity in our main comparisons. The current layer then computes:
\begin{equation}
r_{l,t}
=
\tilde{v}_{l,t}
-
W_{l,t-1}\kappa_{l,t}.
\end{equation}

CLER is a side-channel rather than a new mixer: the recurrent update, gating, and output read are unchanged, and in hybrid stacks the residual is carried through intervening softmax layers to the next routing-capable layer.

\subsection{Cross-Layer Value Routing (CLVR)}
\label{sec:clvr}

\begin{figure}[tbp]
    \centering
    \includegraphics[width=0.85\linewidth]{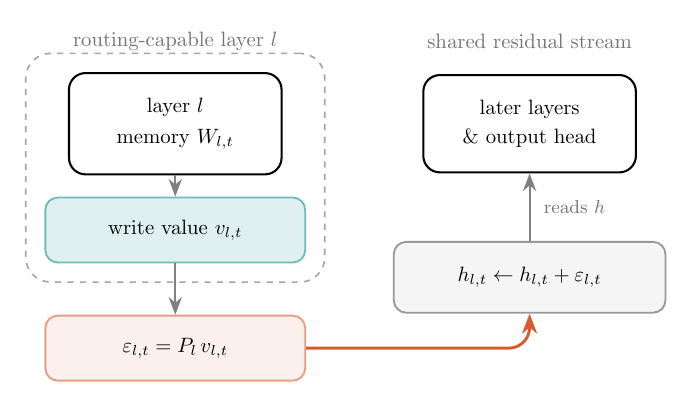}
    \caption{Cross-Layer Value Routing (CLVR). Each routing-capable layer projects its internal write value into the shared residual stream through a zero-initialized projection $P_l$; the routed contribution starts at zero and is added in a depth-aligned space, where it is read by all later layers and the output head.}
    \label{fig:clvr-diagram}
\end{figure}

As reported in Section~\ref{sec:results}, the value-target injection in Eq.~\eqref{eq:cler-value-target} does not improve over matched baselines. We attribute this to a space mismatch: the routed residual stays in the receiver's independently-learned value space, where it is generally misaligned with the higher layer's value geometry and competes with the receiver's own write target. We therefore make two changes: route into the \emph{shared hidden stream} rather than the per-layer value space, and treat the routed quantity as a free choice between the delta-rule error and the write value (\Cref{fig:clvr-diagram}).

Concretely, for a routing-capable layer $l$ we project an internal signal $s_{l,t}$ to the model dimension and add it to the residual stream,
\begin{equation}
\label{eq:clvr-inject}
\varepsilon_{l,t} = P_l\, s_{l,t},
\qquad
h_{l,t} \leftarrow h_{l,t} + \varepsilon_{l,t},
\end{equation}
where $s_{l,t}\in\mathbb{R}^{d_v}$, $h_{l,t}\in\mathbb{R}^{d_{\mathrm{model}}}$ is the ordinary residual-stream state after layer $l$, and $P_l\in\mathbb{R}^{d_{\mathrm{model}}\times d_v}$ is a per-layer projection that is \emph{zero-initialized}, optionally factored as a low-rank product $P_l=U_l D_l$ with $U_l\in\mathbb{R}^{d_{\mathrm{model}}\times d_p}$ and $D_l\in\mathbb{R}^{d_p\times d_v}$. Zero initialization makes the routed contribution vanish at the start of training, so the model begins exactly at the host baseline and learns whether and how strongly to route; the injected $\varepsilon_{l,t}\in\mathbb{R}^{d_{\mathrm{model}}}$ then reaches every later layer and the output head through the standard residual path. Unlike Eq.~\eqref{eq:cler-value-target}, the signal is added in a space that is shared, and therefore aligned, across depth.

We consider two choices for the routed signal $s_{l,t}$. The first is the write \emph{error} $s_{l,t}=r_{l,t}$, which preserves CLER's original motivation of forwarding what a memory failed to store; we call this hidden-stream ablation \emph{CLER-H}. The second is the write \emph{value} $s_{l,t}=v_{l,t}$, the target the memory attempts to store before its own read is subtracted; we call this variant \emph{Cross-Layer Value Routing (CLVR)}. Since $v_{l,t}=r_{l,t}+\bar v_{l,t}$, the two signals differ only by the memory read and occupy the same value space, so $P_l$ has identical shape in both cases and the variants are exactly parameter-matched at every rank. The error-versus-value comparison therefore isolates the \emph{content} of the routed signal rather than its parameter budget.

CLVR is related to, but distinct from, two existing lines of work. Value-residual learning for softmax transformers~\cite{value_residual_learning} adds a fixed early-layer value back into later layers' attention values to counter attention concentration; CLVR instead routes \emph{each} linear-memory layer's own internal write value into the aligned residual stream through a learned, zero-initialized projection, and is defined for delta-rule memories rather than softmax attention. Attention Residuals~\cite{attention_residuals} and related depth-attention schemes route layer \emph{outputs} by a learned softmax over depth and replace the residual sum; CLVR routes a signal taken from \emph{inside} the recurrent operator and adds it rather than replacing the stream. To our knowledge, routing the internal write value of a linear-attention/delta-rule memory into the shared residual stream has not been studied previously.

\section{Experimental Setup}
\label{sec:experiments}

We keep the data, tokenizer, sequence length, global batch size, precision, and hardware fixed across the main 350M comparisons, while allowing architecture-specific layer counts and stack patterns needed to target the 350M-parameter class. Under this setup, differences in validation loss and throughput can be interpreted primarily as architecture, optimizer, or stack-composition effects rather than as changes in the data pipeline. For the main quantitative claims, we require the corresponding run record to identify the model scale, token budget, stack composition, optimizer, learning rate, metric definition, and checkpoint semantics. Exploratory variants with incomplete records are discussed separately as design directions rather than as settled quantitative findings. Appendix~\ref{app:evidence-ledger} summarizes these inclusion criteria for the supplementary material.

\subsection{Shared Training Setup}

Following the experimental setting used by Gated DeltaNet and Gated DeltaNet-2~\cite{yang2024gdn,hatamizadeh2026gdn2}, the experiments use decoder-only language models in the 350M-parameter class, trained on FineWeb-Edu~\cite{penedo2024fineweb} with the LLaMA2 tokenizer~\cite{touvron2023llama2}. The controlled routing ablations use a 20-layer model with hidden size 1024, feed-forward dimension 2816, sequence length 4096, global batch size 128, bf16 precision, and one GH200 node with four GPUs on the CSCS Alps system. The broader architecture comparison keeps this data, tokenizer, sequence length, batch, precision, and hardware recipe fixed while allowing architecture-specific layer counts, evaluating softmax attention, DeltaNet, Gated DeltaNet, Kimi Delta Attention, and Gated DeltaNet-2 under matched training runs up to 15B tokens.

We distinguish between \emph{hybrid} stacks, where linear-attention layers are interleaved with softmax-attention layers, and \emph{pure} stacks, where every mixer layer uses the corresponding linear-attention rule. The default DeltaNet, Gated DeltaNet, and Gated DeltaNet-2 hybrid configurations set one softmax-attention layer every three layers, giving a 2:1 linear-to-softmax pattern.\footnote{For the 20-layer runs, this corresponds to linear-memory mixers at layers $\{0,1,3,4,\ldots,18,19\}$ and softmax-attention mixers at layers $\{2,5,\ldots,17\}$. The 22- and 24-layer stacks follow the same every-third-layer rule. The separate Gated DeltaNet-2 full-attention 3:1 variants set one softmax-attention layer every four layers, giving repeated blocks of three Gated DeltaNet-2 layers followed by one full self-attention layer.} This distinction is central to the empirical results: hybrid stacks often improve loss, while pure stacks expose the throughput advantage of recurrent linear memories.

\begin{table}[tbp]
    \centering
    \caption{Core configuration information for the main experiments. Values are reported from the local launchers and run records when available; entries marked as not fully recorded are not inferred.}
    \label{tab:core-config}
    \footnotesize
    \begin{tabular}{p{0.28\linewidth}p{0.64\linewidth}}
        \toprule
        Item & Configuration \\
        \midrule
        Data and tokenizer & FineWeb-Edu prefix, LLaMA2 tokenizer, train/validation/test split recorded as 99/1/0 in the configurations. \\
        Model scale and depth & Main architecture sweep uses 350M-class models with architecture-specific depths between 20 and 24 layers, chosen to keep total parameter counts close to 312M and within roughly $\pm 2\%$ across variants where recorded. \\
        Controlled routing configuration & Controlled routing ablations use the 20-layer configuration with hidden size 1024, FFN size 2816, 16 attention heads, and 4 query groups. \\
        Sequence and batch & Sequence length 4096, global batch size 128. The 15B-token runs use 16 micro-batch examples per GPU; the compiled PyTorch routing runs use micro-batch 2. \\
        Precision and hardware & bf16 training on one GH200 node with four GPUs for the controlled 350M runs. \\
        Token budgets & Main architecture sweep: 350M-parameter models trained for 15B tokens. Controlled routing ablations: 350M-parameter models trained for 1B tokens. Larger DeltaNet runs: 1.3B-parameter models trained for 40B tokens and 3B-parameter models trained for 60B tokens. \\
        Optimizers & AdamW and Muon/NorMuon. Muon runs use adaptive Muon with NormMuon second-moment handling, spectral scaling, Nesterov momentum 0.95, and scalar learning rate 1.5e-3 where recorded. \\
        Common regularization & Weight decay 0.1, gradient clipping 1.0, Adam betas 0.9/0.95, no attention or hidden dropout in the inspected configurations. \\
        Learning-rate schedule & Warmup-Stable-Decay (WSD) schedule with minus-square-root decay in the main launchers. Recorded 350M 15B defaults are peak LR 3e-4 for AdamW Gated DeltaNet and 3.6e-4 for Muon DeltaNet/Gated DeltaNet configurations; the LR ablation explicitly varies this. \\
        Validation and checkpointing & Validation uses the held-out FineWeb-Edu split. Final-loss entries refer to the recorded final checkpoint/iteration for each run, not the best intermediate checkpoint; best-checkpoint values are used only when a table explicitly says so. \\
        Downstream evaluation & lm-eval-harness via a local-completions API on HellaSwag, PIQA, and WinoGrande; \texttt{acc} is exact-match multiple-choice accuracy, and \texttt{acc\_norm} is the harness's length-normalized likelihood accuracy where produced.\\
        \bottomrule
    \end{tabular}
\end{table}

\paragraph{Metric definitions.}
All validation losses are held-out language-model cross-entropies, so lower is better. In the 15B-token WSD runs, \emph{saturation loss} is the validation loss recorded near the end of the high-learning-rate/saturation phase before the final decay, while \emph{final loss} is the validation loss at the recorded final checkpoint after the full schedule. \emph{Relative speed} is normalized training throughput within the 350M-parameter, 15B-token sweep, with pure Gated DeltaNet under AdamW set to 100\%; it is not an inference-throughput measurement. In the supplementary 1B-token table, \emph{ktok/s/GPU} reports thousands of training tokens processed per second per GPU, without normalization.

\subsection{Optimizer, Learning-Rate, and Iteration-Time Scaling}

The 350M-parameter, 15B-token comparison evaluates AdamW~\cite{loshchilov2019decoupled} and Muon~\cite{jordan2024muon}. The learning-rate ablation trains 350M hybrid models for 2000 steps, or approximately 1.05B tokens, and shows that the best learning rate depends strongly on both optimizer and mixer. In those runs, DeltaNet and Gated DeltaNet with AdamW prefer a learning rate near $10^{-3}$, while the remaining AdamW/Muon combinations cluster around $3 \times 10^{-4}$. We therefore interpret optimizer effects together with learning-rate sensitivity rather than as a single optimizer substitution.

The sequence-length scaling of iteration time is measured at 4k, 16k, and 32k tokens. It is intended to characterize the expected asymptotic behavior of the sequence mixers rather than to replace validation loss as the primary quality metric. Relative speeds in Table~\ref{tab:15b-core-results} are normalized to the fastest 350M-parameter, 15B-token entry, Gated DeltaNet with AdamW in a pure stack.

\subsection{Cross-Layer Routing: Experimental Setup}

Every routing run is compared against a matched non-routing baseline that shares its host architecture, optimizer, schedule, and seed: a routed DeltaNet against DeltaNet, and a routed Gated DeltaNet against Gated DeltaNet. The initial CLER ablations (Table~\ref{tab:cler-results}) use a learned per-layer scalar coefficient $\Gamma$ in Eq.~\eqref{eq:cler-value-target} and raw routed residuals, i.e. $\rho$ is the identity map. Residual-normalized configurations exist as exploratory variants but are not used as main quantitative evidence unless they meet the same matched-run record standard as the reported ablations.

% For the refined routing study of Section~\ref{sec:clvr} we keep the same 350M recipe and add two larger configurations, so that the comparison spans both model and token scale. The matched routing comparisons are run at three scales: 350M parameters / 1B tokens with three seeds, 350M parameters / 15B tokens with one seed, and 1.3B parameters / 40B tokens with two seeds. Each routing run is paired with a non-routing baseline that shares its architecture, optimizer, schedule, and seed, and every comparison is reported separately for the DeltaNet and Gated DeltaNet hosts. Within this study we evaluate three routing pathways under the Muon optimizer: the initial CLER value-target injection (Eq.~\eqref{eq:cler-value-target}), CLER-H ($s_{l,t}=r_{l,t}$ in Eq.~\eqref{eq:clvr-inject}), and CLVR ($s_{l,t}=v_{l,t}$). Because the routing projections are zero-initialized, each routing run starts from exactly its baseline; consistent with the rest of the report we compare the \emph{final} validation loss at the end of training rather than the best intermediate value, since the latter is dominated by evaluation noise. We note that the 15B- and 40B-token runs read from a token-count (\texttt{\_tc}) slice of FineWeb-Edu that differs from the slice used for the 1B-token ablations; absolute losses are therefore comparable only within a fixed token budget and host architecture, and all reported routing gains are differences against the matched baseline at the same scale.

For the refined routing study of Section~\ref{sec:clvr} we keep the same 350M recipe and add a larger Gated DeltaNet configuration, so that the reported comparison spans both model and token scale without implying a complete larger-scale sweep. The matched routing comparisons in Table~\ref{tab:clvr-results} include both DeltaNet and Gated DeltaNet at 350M parameters / 1B tokens and at 350M parameters / 15B tokens. At 1.3B parameters / 40B tokens, the reported hidden-stream routing comparison is available only for the Gated DeltaNet host. Each routing run is paired with a non-routing baseline that shares its architecture, optimizer, schedule, and seed. We did not perform repeated-seed runs for these comparisons, so we report no standard deviations. Within this study we evaluate three routing pathways under the Muon optimizer: the initial CLER value-target injection (Eq.~\eqref{eq:cler-value-target}), CLER-H ($s_{l,t}=r_{l,t}$ in Eq.~\eqref{eq:clvr-inject}), and CLVR ($s_{l,t}=v_{l,t}$). Because the routing projections are zero-initialized, each routing run starts from exactly its baseline; consistent with the rest of the report we compare the \emph{final} validation loss at the end of training rather than the best intermediate value, since the latter is dominated by evaluation noise. We note that the 15B- and 40B-token runs read from a token-count (\texttt{\_tc}) slice of FineWeb-Edu that differs from the slice used for the 1B-token ablations; absolute losses are therefore comparable only within a fixed token budget and host architecture, and all reported routing gains are differences against the matched baseline at the same scale.

We evaluate cross-layer routing only on DeltaNet and Gated DeltaNet hosts, so no empirical claim is made about these routing variants on top of Kimi Delta Attention or Gated DeltaNet-2.

\section{Results}
\label{sec:results}

This section reports the empirical findings in increasing order of scope: 350M-parameter models trained for 15B tokens, optimizer and learning-rate behavior, sequence-length scaling of training iteration time, larger DeltaNet runs, downstream evaluations, and finally validation and residual analysis for cross-layer routing.

All reported architecture and routing comparisons are single-run comparisons; we did not perform repeated-seed runs and therefore do not report standard deviations. We therefore abstain from drawing strong conclusions from small validation-loss differences, especially when differences are below roughly $10^{-3}$ to $10^{-2}$ and are not supported by matched settings, consistent optimizer trends, or downstream agreement. Larger effects, such as the 32k iteration-time gap between softmax attention and pure recurrent stacks, the consistent Muon improvements in matched 350M-parameter, 15B-token rows, and the best final-loss entry in Table~\ref{tab:15b-core-results}, remain meaningful within the reported configuration.

\subsection{350M-Parameter, 15B-Token Validation-Loss/Throughput Frontier}

\begin{figure}[tbp]
    \centering
    \includegraphics[width=\linewidth]{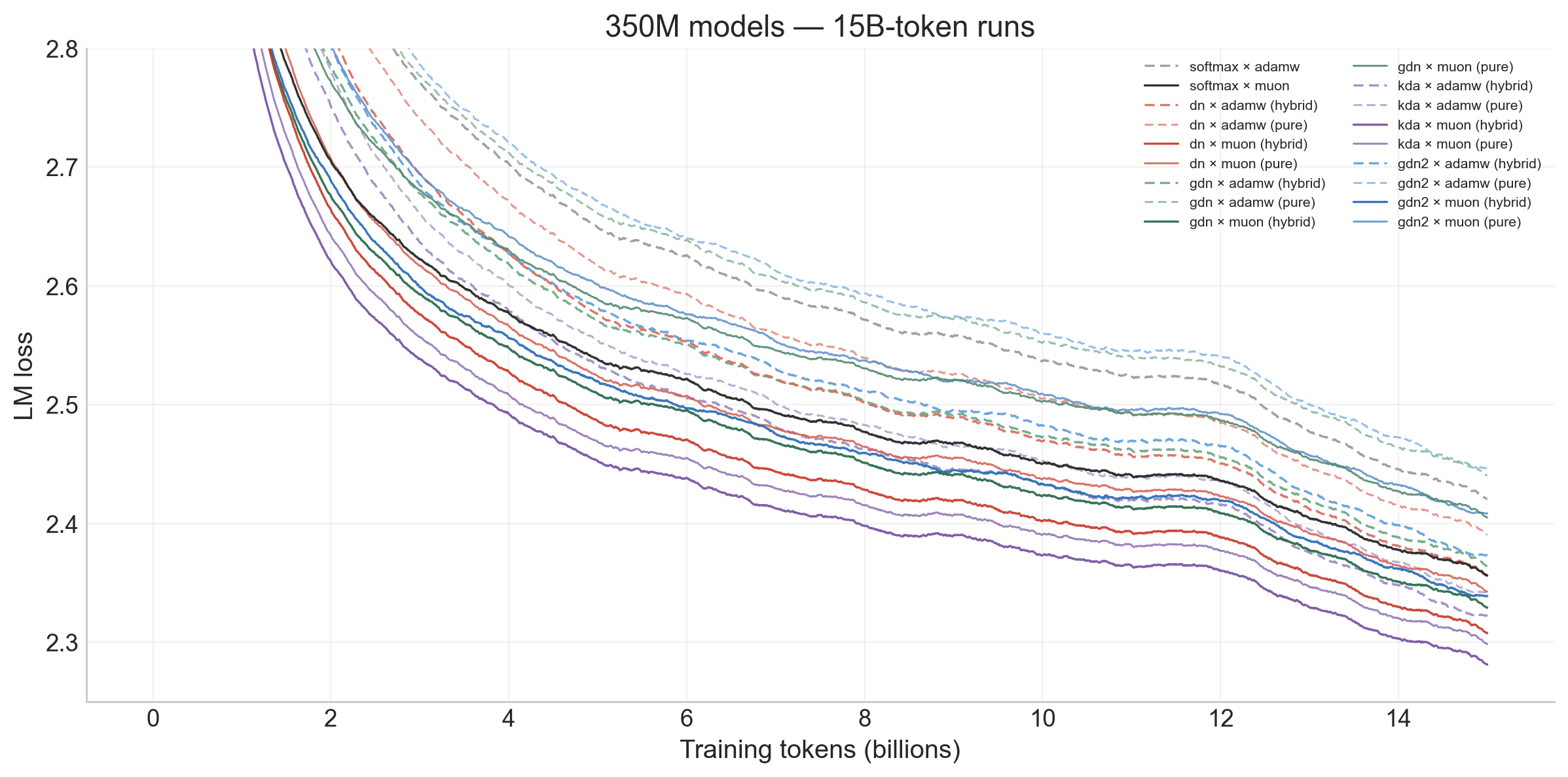}
    \caption{Validation-loss scaling for the 350M-parameter runs trained to 15B tokens. Lower is better. In this sweep, Kimi Delta Attention with Muon in a hybrid stack reaches the lowest final loss, while DeltaNet and Gated DeltaNet remain competitive with simpler recurrent-memory updates.}
    \label{fig:15b-scaling-loss}
\end{figure}

Figure~\ref{fig:15b-scaling-loss} and Table~\ref{tab:15b-core-results} summarize the 350M-parameter runs trained for 15B tokens. The best final validation loss among these reported runs is Kimi Delta Attention with Muon in a hybrid stack, which reaches 2.273. DeltaNet with Muon in a hybrid stack is the strongest non-KDA entry at 2.299, while Gated DeltaNet-2 with Muon in a hybrid stack reaches 2.345. The fastest entry is pure Gated DeltaNet with AdamW, normalized to 100\% relative speed, but this setting has a substantially higher final loss of 2.433 than the best hybrid Muon runs.

Two patterns are consistent across the table. First, Muon improves the matched final loss for every architecture family shown here. Second, hybrid stacks usually improve loss relative to pure stacks, but they reduce part of the speed advantage that motivates linear attention. The resulting trade-off is visible in the reported sweep: the lowest loss comes from a slower KDA hybrid run, while the fastest pure recurrent stack is not the strongest model by validation loss.

\begin{table}[tbp]
    \centering
    \caption{350M-parameter, 15B-token architecture comparison. Saturation loss is measured before the final WSD decay, final loss at the final recorded checkpoint, and relative speed is normalized training throughput against the fastest entry in this sweep, pure Gated DeltaNet with AdamW. Lower losses are better; higher relative speed is faster. Values are directly comparable within this sweep, but each row is a single run rather than a repeated-seed estimate.}
    \label{tab:15b-core-results}
    \resizebox{\linewidth}{!}{\begin{tabular}{lllrrr}
\toprule
Architecture & Optimizer & Stack & Relative speed & Saturation loss & Final loss \\
\midrule
Softmax & AdamW & dense & 81.7\% & 2.587 & 2.413 \\
Softmax & Muon & dense & 79.7\% & 2.489 & 2.349 \\
DeltaNet & AdamW & hybrid & 87.0\% & 2.517 & 2.348 \\
DeltaNet & AdamW & pure & 91.9\% & 2.555 & 2.382 \\
DeltaNet & Muon & hybrid & 83.8\% & 2.440 & 2.299 \\
DeltaNet & Muon & pure & 89.2\% & 2.476 & 2.334 \\
Gated DeltaNet & AdamW & hybrid & 90.5\% & 2.517 & 2.356 \\
Gated DeltaNet & AdamW & pure & \textbf{100.0\%} & 2.601 & 2.433 \\
Gated DeltaNet & Muon & hybrid & 89.5\% & 2.464 & 2.321 \\
Gated DeltaNet & Muon & pure & 95.0\% & 2.543 & 2.397 \\
Kimi Delta Attention & AdamW & hybrid & 74.4\% & 2.471 & 2.328 \\
Kimi Delta Attention & AdamW & pure & 73.6\% & 2.490 & 2.347 \\
Kimi Delta Attention & Muon & hybrid & 70.9\% & \textbf{2.409} & \textbf{2.273} \\
Kimi Delta Attention & Muon & pure & 68.5\% & 2.427 & 2.290 \\
Gated DeltaNet-2 & AdamW & hybrid & 83.3\% & 2.520 & 2.379 \\
Gated DeltaNet-2 & AdamW & pure & 81.9\% & 2.601 & 2.452 \\
Gated DeltaNet-2 & Muon & hybrid & 81.3\% & 2.466 & 2.345 \\
Gated DeltaNet-2 & Muon & pure & 80.7\% & 2.542 & 2.415 \\
\bottomrule
\end{tabular}
}
\end{table}

\subsection{Learning-Rate and Optimizer Effects}

\begin{figure}[tbp]
    \centering
    \includegraphics[width=\linewidth]{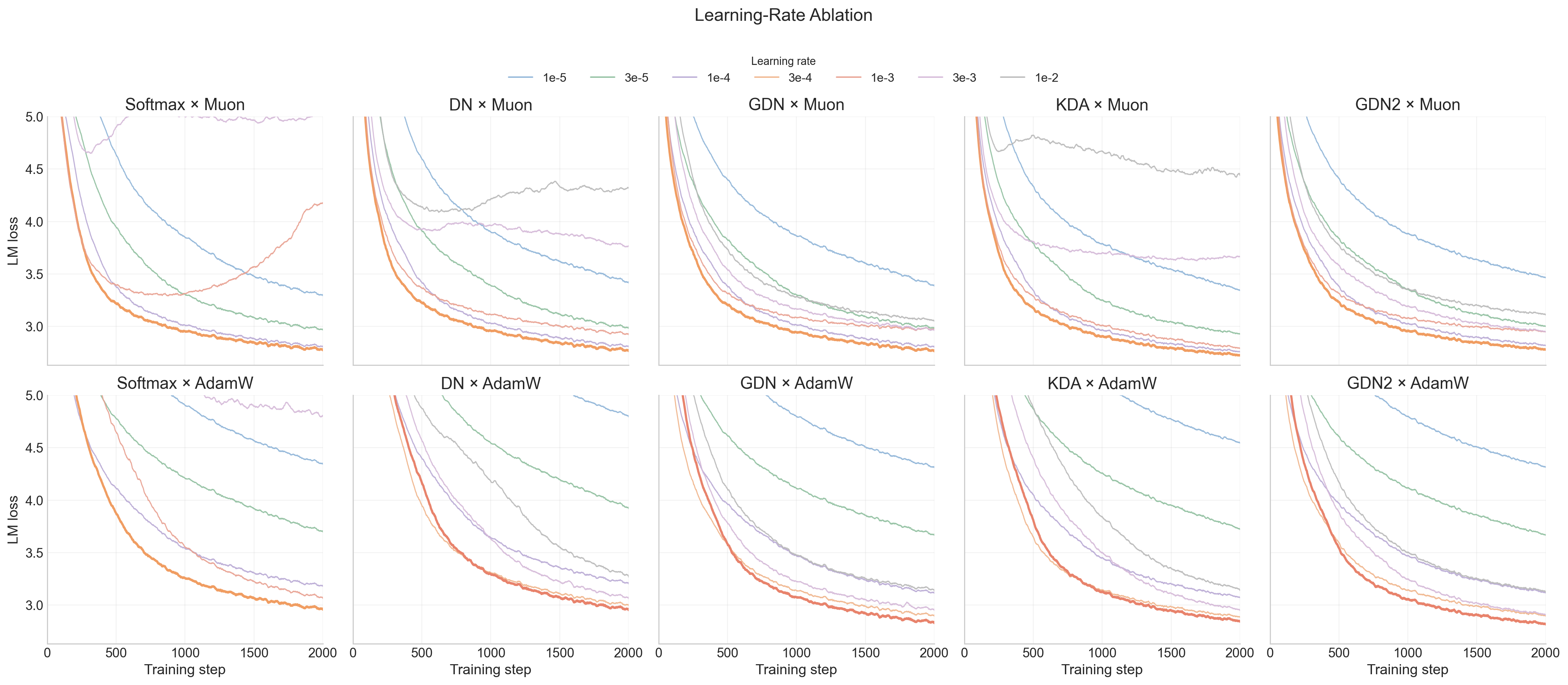}
    \caption{Learning-rate ablation for 350M-parameter hybrid models trained for 2000 steps, approximately 1.05B tokens. The preferred learning rate varies by optimizer and mixer, making learning-rate tuning necessary before drawing architectural conclusions.}
    \label{fig:lr-ablation}
\end{figure}

\begin{figure}[tbp]
    \centering
    \includegraphics[width=0.6\linewidth]{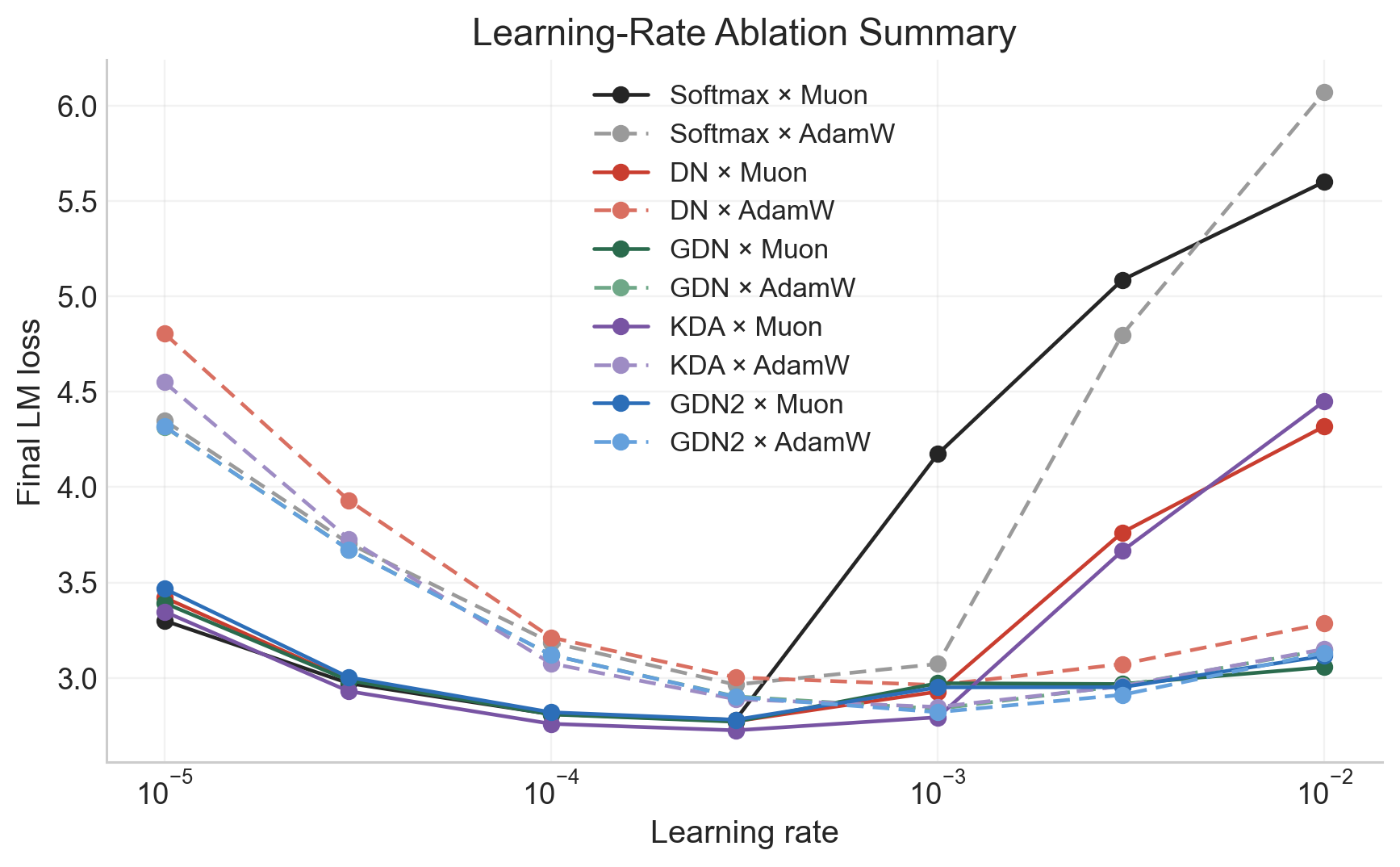}
    \caption{Learning-rate ablation iso plots for 350M-parameter hybrid models trained for 2000 steps. The iso plots show the final loss interpolation line for different learning rates. The loss increasing on the left and right ends for all attention/optimizer combination implies the covered range is sufficient, reaching from a too-low learning rate over the optimum up to a too-high learning rate. The optimum lies between $3 \times 10^{-4} $ and $10^{-3}$ depending on attention/optimizer.}
    \label{fig:lr-ablation-iso}
\end{figure}

Figure~\ref{fig:lr-ablation} shows that the optimizer comparison is not separable from learning-rate choice. At the 350M-parameter hybrid setting, Muon prefers a lower learning rate around $3 \times 10^{-4}$ (with $10^{-4}$ being close/catching up), whereas all AdamW linear attention variants prefer $10^{-3}$. The exception is standard softmax attention, which matches its Muon counterpart with $3 \times 10^{-4}$.
This shows that a single default learning rate can distort architecture comparisons: a variant may underperform because its optimizer/learning-rate pair is poorly matched, rather than because the recurrent rule itself is inferior. The iso plots in \Cref{fig:lr-ablation-iso} visualize the same runs, showing the final-loss interpolation line for different learning rates. It demonstrates the plateau length in which performance is close to optimal: For Muon (the solid lines), the plateau is slightly longer and on lower learning rates. Additionally, softmax is the most dependent on a good learning rate choice, whereas linear attention architectures are more forgiving, especially on higher learning rates.

These results indicate that Muon should be treated as part of the architecture evaluation protocol rather than as an optimizer detail independent of the model family. In Table~\ref{tab:15b-core-results}, Muon improves final loss for every matched architecture/stack pair, but the magnitude of the improvement varies across softmax attention, DeltaNet, Gated DeltaNet, Kimi Delta Attention, and Gated DeltaNet-2.

\subsection{Sequence-Length Iteration-Time Scaling}

\begin{figure}[tbp]
    \centering
    \includegraphics[width=0.85\linewidth]{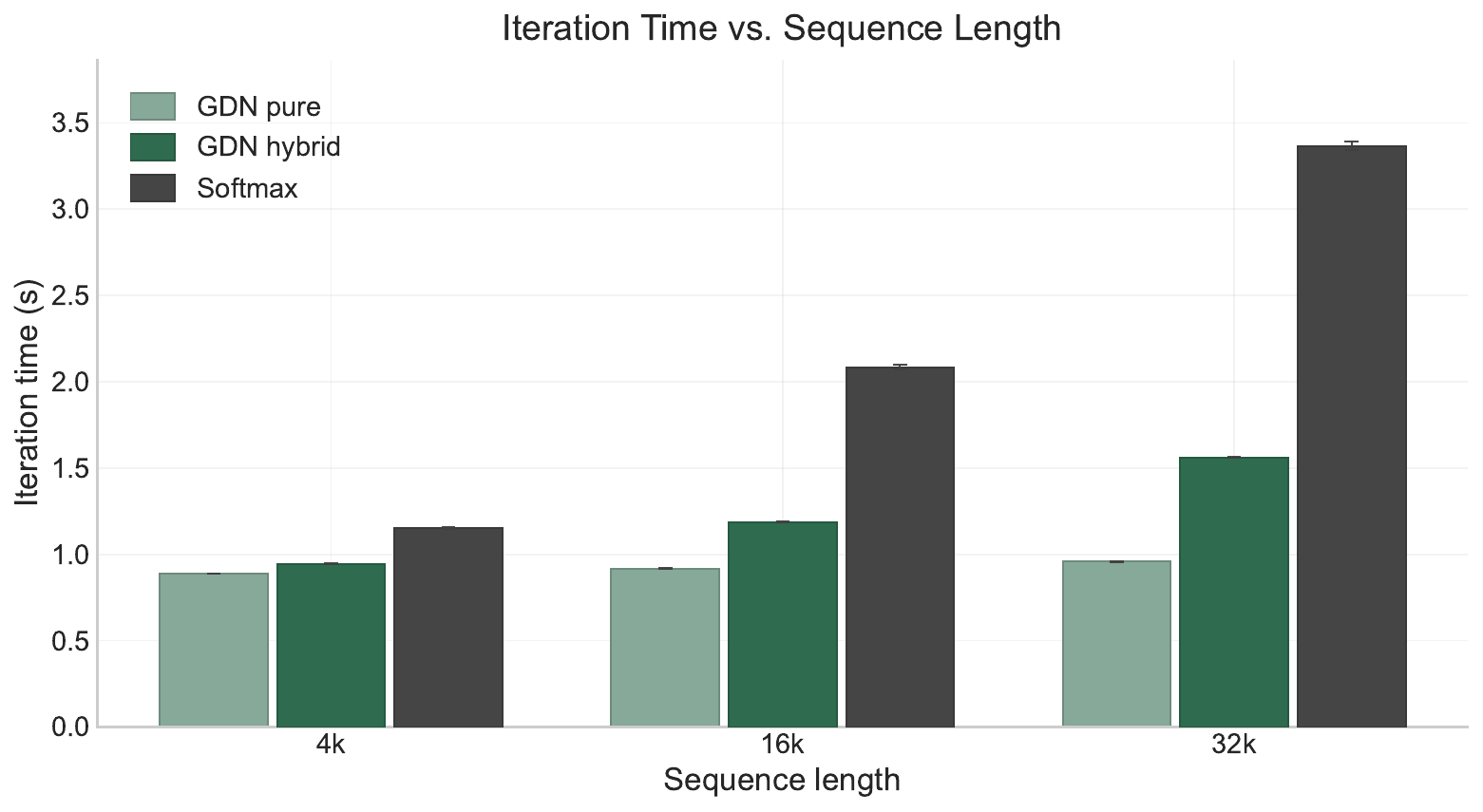}
    \caption{Iteration-time scaling from 4k to 32k tokens. Softmax attention time grows much faster with sequence length ($+192\%$) than Gated DeltaNet hybrid ($+65\%$) and pure stacks ($+8\%$). Note that hybrid iteration time increase is very close to $1/3$ of softmax, matching the fact that hybrid has $1/3$ softmax layers. Error bars (step-to-step variance) are very low.}
    \label{fig:iteration-time}
\end{figure}

Figure~\ref{fig:iteration-time} measures iteration time as the sequence length increases from 4k to 32k tokens. At 32k tokens, the timing measurement is 3.37 seconds per iteration for softmax attention, 1.56 seconds for a Gated DeltaNet hybrid stack, and 0.96 seconds for a pure Gated DeltaNet stack. From 4k to 32k tokens, the corresponding growth factors are approximately $2.9\times$, $1.7\times$, and $1.1\times$. This provides direct empirical evidence for the expected scaling advantage of recurrent linear-memory mixers, especially when the stack is pure.

The timing result also clarifies the cost of hybridization. Hybrid stacks often recover validation loss relative to pure linear stacks, but the retained softmax layers reintroduce part of the sequence-length cost. The preferred stack therefore depends on whether the target setting prioritizes lower loss at moderate context length or flatter iteration-time scaling at longer context length.

\subsection{Larger DeltaNet Runs and Downstream Evaluation}

Table~\ref{tab:large-deltanet-runs} extends the DeltaNet comparison beyond the 350M-parameter scale. At 1.3B parameters and 40B tokens, the best listed final loss is 2.063 for a pure DeltaNet run at learning rate $1.5 \times 10^{-4}$, while the corresponding pure CLER-DeltaNet run at the same learning rate reaches 2.112. At 3B parameters and 60B tokens, the two better DeltaNet hybrid runs reach 1.955 at learning rates $3 \times 10^{-4}$ and $1.5 \times 10^{-4}$, while the run at $5 \times 10^{-4}$ has a substantially higher final loss of 2.332. These runs reinforce the learning-rate sensitivity seen at smaller scale.

\begin{table}[tbp]
    \centering
    \caption{Larger DeltaNet-only runs at 1.3B-parameter/40B-token and 3B-parameter/60B-token scale. Lower final loss is better. These rows illustrate larger-scale DeltaNet sensitivity to learning rate and stack pattern, but they are not a full architecture sweep across all mechanisms.}
    \label{tab:large-deltanet-runs}
    \begin{tabular}{lllr}
\toprule
Run & Learning rate & Scale & Final loss \\
\midrule
\texttt{dn-hybrid} & 3.6e-4 & 1.3B/40B & 2.066 \\
\texttt{dn-pure} & 3.6e-4 & 1.3B/40B & 2.085 \\
\texttt{dn-hybrid} & 1.5e-4 & 1.3B/40B & 2.109 \\
\texttt{dn-pure} & 1.5e-4 & 1.3B/40B & \textbf{2.063} \\
\texttt{dn-pure-cler} & 1.5e-4 & 1.3B/40B & 2.112 \\
\midrule
\texttt{dn-hybrid} & 5e-4 & 3B/60B & 2.332 \\
\texttt{dn-hybrid} & 3e-4 & 3B/60B & \textbf{1.955} \\
\texttt{dn-hybrid} & 1.5e-4 & 3B/60B & \textbf{1.955} \\
\bottomrule
\end{tabular}

\end{table}

%% TODO: check if current new phrasing is ok (rephrased so that winogrande is not used as a clear signal)
% Table~\ref{tab:downstream} reports downstream checks on HellaSwag, PIQA, and WinoGrande. We use them as supporting evidence about whether the checkpoints remain broadly usable, not as a decisive benchmark of downstream superiority. The 350M-parameter results vary across tasks. At 1.3B parameters and 40B tokens, the DeltaNet hybrid model has the highest listed HellaSwag and PIQA scores among the 1.3B entries. The CLER-DeltaNet pure variants in this 1.3B block use the earlier $\gamma$ value-target formulation rather than the hidden-stream routing variants; they do not improve HellaSwag or PIQA relative to the pure no-CLER baseline, while the $\gamma=0.0$ row is slightly higher on WinoGrande. The 3B-parameter, 60B-token DeltaNet hybrid row improves substantially over the 1.3B rows, especially on HellaSwag normalized accuracy.

\begin{table}[tbp]
    \centering
    \caption{Downstream checks on HellaSwag, PIQA, and WinoGrande using the local lm-eval-harness API path. Accuracy is exact-match multiple-choice accuracy; normalized accuracy is the harness's length-normalized likelihood accuracy where produced.}
    \label{tab:downstream}
    \resizebox{\linewidth}{!}{\begin{tabular}{llllrr|rr|rr|r}
\toprule
\multicolumn{6}{c|}{} & \multicolumn{2}{c|}{HellaSwag} & \multicolumn{2}{c|}{PIQA} & WinoGrande \\
\cmidrule(lr){7-8}\cmidrule(lr){9-10}\cmidrule(lr){11-11}
Size & Arch. & Variant & Optim. & Iter. & Tokens & acc & acc\_norm & acc & acc\_norm & acc \\
\midrule
350M & GDN & hybrid & Muon & 28610 & 15.0B & 0.3480 & 0.4158 & 0.6676 & 0.6741 & \textbf{0.5414} \\
350M & GDN & pure & Muon & 28610 & 15.0B & 0.3337 & 0.3978 & \textbf{0.6763} & 0.6741 & 0.4870 \\
350M & DeltaNet & hybrid & Muon & 28610 & 15.0B & \textbf{0.3541} & \textbf{0.4305} & 0.6681 & 0.6779 & 0.5051 \\
350M & DeltaNet & pure & Muon & 28610 & 15.0B & 0.3426 & 0.4133 & 0.6741 & \textbf{0.6801} & 0.5170 \\
350M & GDN & hybrid & AdamW & 28610 & 15.0B & 0.3369 & 0.3992 & 0.6616 & 0.6616 & 0.5099 \\
350M & GDN & pure & AdamW & 28610 & 15.0B & 0.3265 & 0.3817 & 0.6561 & 0.6572 & 0.5217 \\
350M & DeltaNet & hybrid & AdamW & 28610 & 15.0B & 0.3361 & 0.3987 & 0.6654 & 0.6513 & 0.5249 \\
350M & DeltaNet & pure & AdamW & 28610 & 15.0B & 0.3293 & 0.3918 & 0.6480 & 0.6551 & 0.5154 \\
\midrule
1.3B & DeltaNet & hybrid & Muon & 19073 & 40.0B & \textbf{0.4277} & \textbf{0.5484} & \textbf{0.7209} & \textbf{0.7301} & 0.5572 \\
1.3B & DeltaNet & pure, no CLER & Muon & 19073 & 40.0B & 0.4119 & 0.5234 & 0.7106 & 0.7193 & 0.5612 \\
1.3B & CLER-DeltaNet & pure, $\gamma{=}0.1$ & Muon & 19073 & 40.0B & 0.4049 & 0.5231 & 0.7089 & 0.7203 & 0.5446 \\
1.3B & CLER-DeltaNet & pure, $\gamma{=}0.0$ & Muon & 19073 & 40.0B & 0.4078 & 0.5220 & 0.7095 & 0.7160 & \textbf{0.5627} \\
\midrule
3B & DeltaNet & hybrid & Muon & 7152 & 60.0B & 0.4617 & 0.6063 & 0.7334 & 0.7410 & 0.5848 \\
\bottomrule
\end{tabular}
}
\end{table}

Table~\ref{tab:cler-downstream-15b} reports the matched 350M-parameter, 15B-token Muon downstream evaluations for the CLER-H and CLVR checkpoints. The routing variants are broadly comparable to their matched Gated DeltaNet and DeltaNet baselines on HellaSwag and PIQA, with small mixed differences across accuracy and normalized accuracy. Although the WinoGrande point estimates are higher for the routed variants, these task-specific single-checkpoint differences are not sufficiently reliable to serve as evidence of a routing effect. We therefore use the downstream evaluations only as a check for clear degradation, which we do not observe on the evaluated tasks.

\begin{table}[tbp]
    \centering
    \caption{Matched 350M-parameter, 15B-token Muon downstream evaluations for Gated DeltaNet and DeltaNet routing checkpoints. All rows use the scoring-server lm-eval-harness path with full-sequence forward passes, the LLaMA2 tokenizer, batch size 8, maximum length 4096, and zero-shot evaluation.}
    \label{tab:cler-downstream-15b}
    \resizebox{\linewidth}{!}{\begin{tabular}{llrrrrr}
\toprule
Host & Variant & HellaSwag acc & HellaSwag acc\_norm & PIQA acc & PIQA acc\_norm & WinoGrande acc \\
\midrule
Gated DeltaNet & Baseline & \textbf{0.3434} & \textbf{0.4131} & 0.6692 & \textbf{0.6649} & 0.5146 \\
Gated DeltaNet & Route error (CLER-H) & 0.3423 & 0.4104 & 0.6670 & 0.6556 & 0.5209 \\
Gated DeltaNet & Route value (CLER-V/CLVR) & 0.3421 & 0.4113 & \textbf{0.6719} & 0.6600 & \textbf{0.5328} \\
\midrule
DeltaNet & Baseline & 0.3410 & \textbf{0.4113} & 0.6600 & 0.6605 & 0.5193 \\
DeltaNet & Route error (CLER-H) & 0.3424 & 0.4069 & \textbf{0.6681} & \textbf{0.6757} & 0.5264 \\
DeltaNet & Route value (CLER-V/CLVR) & \textbf{0.3437} & 0.4104 & 0.6616 & 0.6551 & \textbf{0.5359} \\
\bottomrule
\end{tabular}
}
\end{table}

\subsection{CLER Validation Loss}

Table~\ref{tab:cler-results} summarizes the matched CLER comparisons at the 350M-parameter, 1B-token scale. Under Muon, CLER-Gated has slightly higher final loss than Gated DeltaNet, while CLER-DeltaNet obtains a very small improvement over DeltaNet; the latter difference is below $10^{-3}$ and comes from a single run, so we do not interpret it as a robust gain. Under AdamW, CLER does not improve either host. The pathway is not inactive (the routed residuals are nonzero at every routing-capable layer), but the matched trajectories are nearly overlapping, consistent with the space-mismatch argument of Section~\ref{sec:clvr}: the error routed into the per-layer value target appears either insufficiently influential or poorly aligned with the receiver's learned dynamics.

\begin{table}[tbp]
    \centering
    \caption{Matched CLER comparisons at 350M parameters and approximately 1B training tokens. $\Delta$ is computed relative to the corresponding non-CLER baseline under the same optimizer; negative values indicate lower final validation loss. The small Muon CLER-DeltaNet gain is below $10^{-3}$ and is not interpreted as a robust improvement.}
    \label{tab:cler-results}
    \begin{tabular}{llcc}
        \toprule
        Optimizer & Variant & Final val loss & $\Delta$ vs. baseline \\
        \midrule
        Muon & Gated DeltaNet & 2.8319 & -- \\
        Muon & CLER-Gated, scalar & 2.8333 & $+0.0013$ \\
        Muon & DeltaNet & 2.8511 & -- \\
        Muon & CLER-DeltaNet, scalar & 2.8507 & $-0.0004$ \\
        \midrule
        AdamW & Gated DeltaNet & 3.2554 & -- \\
        AdamW & CLER-Gated, scalar & 3.2562 & $+0.0008$ \\
        AdamW & DeltaNet & 3.4068 & -- \\
        AdamW & CLER-DeltaNet, scalar & 3.4258 & $+0.0190$ \\
        \bottomrule
    \end{tabular}
\end{table}

\subsection{Cross-Layer Value Routing Results}
\label{sec:clvr-results}

The negative result above is consistent with the original injection adding a small, and possibly misaligned, signal in the wrong space. Section~\ref{sec:clvr} addresses both concerns at once: it routes into the shared hidden stream through a zero-initialized projection (Eq.~\eqref{eq:clvr-inject}) rather than into the per-layer value target, and it treats the routed quantity as a choice between the write error and the write value. Table~\ref{tab:clvr-results} reports the resulting matched comparisons for the available host/scale settings.

\begin{table}[tbp]
    \centering
    \caption{Matched hidden-stream routing comparisons under Muon. CLER-H routes the write residual ($s_{l,t}=r_{l,t}$), while CLVR routes the write value ($s_{l,t}=v_{l,t}$) in Eq.~\eqref{eq:clvr-inject}. Baseline denotes the final validation loss of the matched non-routing model, and $\Delta$ is measured relative to that baseline at the same host and scale; negative values indicate lower loss. All entries are single matched runs; no standard deviations are available. The 1.3B/40B hidden-stream routing comparison is reported for Gated DeltaNet only. Absolute losses are comparable only within a row because the token-count slices differ across scales (Section~\ref{sec:experiments}).}

    \label{tab:clvr-results}
    \begin{tabular}{llcrr}
        \toprule
        Host & Params / tokens & Baseline & CLER-H $\Delta$ & CLVR $\Delta$ \\
        \midrule
        Gated DeltaNet & 350M / 1B  & 2.8331 & $-0.0073$ & $\mathbf{-0.0103}$ \\
        Gated DeltaNet & 350M / 15B & 2.3417 & $-0.0042$ & $\mathbf{-0.0059}$ \\
        Gated DeltaNet & 1.3B / 40B & 2.0635 & $-0.0010$ & $\mathbf{-0.0019}$ \\
        DeltaNet       & 350M / 1B  & 2.8469 & $-0.0047$ & $\mathbf{-0.0119}$ \\
        DeltaNet       & 350M / 15B & 2.3347 & $-0.0002$ & $\mathbf{-0.0016}$ \\
        \bottomrule
    \end{tabular}
\end{table}

Two findings are consistent across every reported row. First, moving the routing target from the per-layer value space, which does not improve over baseline in Table~\ref{tab:cler-results}, to the aligned hidden stream turns the comparison from neutral-or-negative into a small positive effect. Second, CLVR is uniformly better than CLER-H, even though the two signals occupy the same space and use an identically shaped projection. Because the two variants differ only by the receiver's own memory read, this comparison isolates the routed \emph{content} as the useful ingredient and shows that the delta-rule error, the signal that motivated CLER, is not the right quantity to forward. Since $v_{l,t}=r_{l,t}+\bar v_{l,t}$, the write error is the write value minus the memory's own read, so routing the error still carries the part of the value that the lower memory has not yet absorbed; this is consistent with error routing helping, but consistently less than routing the full value.

The gains are small and point in the same direction across the reported single-run rows. At 350M parameters and 1B tokens CLVR lowers final validation loss by about $0.010$ to $0.012$ on \emph{both} hosts ($-0.0103$ for Gated DeltaNet, $-0.0119$ for DeltaNet). The gain then shrinks in the longer 350M/15B runs on both hosts and in the available larger Gated DeltaNet row, reaching $-0.0059$ and $-0.0016$ at 350M/15B and $-0.0019$ at 1.3B/40B. It does not reverse in any reported row, and the value-over-error ordering is preserved throughout Table~\ref{tab:clvr-results}; however, the table does not establish larger-scale behavior for DeltaNet because the 1.3B/40B DeltaNet hidden-stream routing comparison is not reported. This pattern is consistent with cross-layer routing supplying information that a smaller or less-trained recurrent memory has not yet captured on its own, with diminishing headroom as the host is strengthened. For the runs where we log it, the zero-initialized projection $P_l$ grows to a comparable, nonzero norm across all routing-capable layers, confirming that the trained model actively uses the pathway rather than leaving it a negligible perturbation.

We also tested two natural extensions of value routing at 350M/1B (single seed): gating the routed value by a local surprise signal $\|r_{l,t}\|/\|v_{l,t}\|$, and routing the concatenation $[r_{l,t};v_{l,t}]$ of error and value. Neither improved on plain value routing with final losses of $2.8220$ and $2.8194$ respectively, against $2.8152$ for CLVR on the same seed with the surprise gate being slightly worse. These controls are single-seed and their differences are within run-to-run noise, but they reinforce the main finding that the unmodified write value is the signal worth routing, and that re-introducing the error or modulating the value does not help. Two further single-seed controls attribute the gain to the routed signal rather than to the added parameters. Routing the layer's own hidden state through the same zero-initialized projection, instead of an internal mixer signal, is essentially flat ($\Delta\approx-0.0014$), and enlarging the Gated DeltaNet baseline by the same parameter count does not help either ($+0.0021$); both fall well short of the routing gains, so the improvement comes from \emph{what} is routed rather than from the extra capacity.

For context, we also evaluated a depth-attention baseline, Attention Residuals~\cite{attention_residuals}, which combines sub-layer outputs through a learned softmax over depth rather than additively routing an internal signal. Its effect is host-dependent in our 350M/15B runs: it improves DeltaNet (final loss $2.3246$, $-0.0101$ versus the DeltaNet baseline) but slightly worsens Gated DeltaNet ($+0.0030$). CLVR is more uniform across the two hosts in these runs, though its gains are smaller than the DeltaNet-only Attention Residuals gain. We do not pursue Attention Residuals further here, because it replaces the residual sum and routes layer outputs rather than an internal write value, and because its benefit does not transfer across the two linear-memory hosts we study.

\subsection{Reproducibility}

The implementation and experiment records are available at \githubrepo. The released repository is a focused Megatron-LM fork containing the linear-attention implementations, final training launchers, FineWeb-Edu/LLaMA2 data-preparation path, smoke checks, and the scoring-server evaluation wrapper used for the downstream results. For reproduction, the launchers are the executable source of truth: they pin the architecture, optimizer, token budget, checkpoint format, and evaluation cadence, while site-specific paths such as the tokenizer, Megatron data prefix, W\&B credentials, and output directories are supplied through environment variables. Table~\ref{tab:core-config} summarizes the shared configuration, and the run-specific scripts provide the remaining details needed to rerun the 350M, larger-scale, and routing comparisons. Exploratory routing controls should be treated as reproducible paper claims only when they are accompanied by complete run records with the same metric definitions and checkpoint semantics used for the main tables.

\section{Discussion: Validation Loss, Throughput, and Architectural Trade-offs}
\label{sec:discussion}

The common notation separates the main design choices in the DeltaNet family: how the memory is read, how old content is forgotten or erased, and how new value information is written. The empirical results show that these choices should not be interpreted as a one-dimensional architecture ranking. They instead define a multi-objective frontier involving validation loss, training throughput, sequence-length scaling, and implementation complexity.

In this frontier, no single setting dominates every objective. Kimi Delta Attention with Muon reaches the best 350M-parameter, 15B-token validation loss in our experiments, but it is also among the slower entries. Pure recurrent stacks give the strongest long-context iteration-time scaling, while hybrid stacks usually recover part of the validation-loss gap at the cost of some throughput. The practical question is therefore not simply whether to use softmax attention or linear attention, but which combination of mixer, stack pattern, optimizer, and learning rate best matches the target operating point.

\paragraph{Evidence limitations.}
The empirical comparisons are an audited run set rather than a full statistical study. The reported architecture and routing rows are single runs, and the available records do not support seed-averaged rankings or standard deviations. Hyperparameter coverage is also uneven across variants, so optimizer and learning-rate effects should be read together. The downstream evaluation is limited to HellaSwag, PIQA, and WinoGrande, and may miss behaviors that matter for long-context or memory-intensive use cases. Appendix~\ref{app:evidence-ledger} states the resulting inclusion rule: incomplete branch records are treated as exploratory design evidence, not as main numerical claims.

%% TODO: check, I (lingfeng) changed supportive evidence to compatibility checks
The cross-layer routing results are mixed but coherent. In its DeltaRule-inspired formulation, where a lower layer's delta-rule write error is injected into the next layer's value target (Eq.~\eqref{eq:cler-value-target}), CLER is active but does not yield a reliable validation-loss improvement in the controlled settings reported here: routed residuals are nonzero and the receiver-side coefficients are trainable, yet the matched comparisons are flat (Table~\ref{tab:cler-results}). We trace this to a basis mismatch. The delta-rule residual is a local correction defined by one layer's memory state, key representation, value target, and feature map; once routed into another layer's independently learned value space it need not be aligned, and it competes directly with the receiver's own write target. Two adaptations resolve the issue (Section~\ref{sec:clvr}): routing into the shared, and therefore aligned, residual stream rather than the per-layer value space, and forwarding the write \emph{value} rather than the write error. The resulting method, CLVR, gives a small improvement over the reported matched DeltaNet and Gated DeltaNet baselines (Table~\ref{tab:clvr-results}), with the larger-scale hidden-stream routing comparison available for Gated DeltaNet only. The matched downstream evaluations are more limited: HellaSwag and PIQA are broadly comparable to the baselines, while WinoGrande moves in the favorable direction for both CLER-H and CLVR (Table~\ref{tab:cler-downstream-15b}), these differences are not reliable enough here to serve as an independent signal. We interpret these evaluations as showing no clear downstream degradation on the tested tasks, not as proof of broad downstream improvement. That value beats error at fixed parameter count indicates the useful cross-layer signal is the layer's write target, not the delta-rule correction that originally motivated CLER. The choice is left open for richer hosts such as Gated DeltaNet-2, where the write value and the separate erase and write residuals are decoupled.

\section{Future Work}

Although CLVR helps in the reported DeltaNet and Gated DeltaNet rows, the effect is small and leaves several questions open. The clearest is whether the gain persists, or grows, on hosts we did not evaluate under routing: Kimi Delta Attention and Gated DeltaNet-2. Gated DeltaNet-2 in particular decouples the write value from separate erase and write residuals, so it is not obvious which internal signal should be routed; a matched comparison there would test whether ``route the write value'' generalizes beyond DeltaNet and Gated DeltaNet. It also remains open whether the diminishing return we observe with longer training and in the larger Gated DeltaNet row continues until CLVR becomes neutral, or settles at a small positive floor, which additional repeated runs at larger token budgets could resolve.

Future routing runs should also log the quantities needed to measure influence directly: the routed signal after projection, the receiver's residual stream before and after injection, and the learned projection norms at matched checkpoints. These logs would make it possible to separate cases that validation loss alone cannot distinguish, namely whether a remaining gap is because the routed signal is too small, expressed in the wrong basis, or aligned but redundant with the receiver's local update.

Finally, CLVR should be tested on tasks where cross-layer recovery has a clearer role than average next-token prediction. Synthetic key-value retrieval, passkey-style tasks, long-context QA, many-distractor in-context learning, and other associative-recall settings are natural candidates. The throughput study should also be extended from training iteration time to measured inference throughput, decoding memory footprint, and quality at long context length.

\section{Conclusion}

We compared softmax attention with DeltaNet, Gated DeltaNet, Kimi Delta Attention, and Gated DeltaNet-2 in a common recurrent-memory notation. The empirical picture is a multi-objective trade-off among validation loss, throughput, and sequence-length scaling rather than a single architecture ranking.

Within the reported 350M-parameter, 15B-token sweep, Kimi Delta Attention with Muon in a hybrid stack reaches the best validation loss, while pure Gated DeltaNet with AdamW gives the fastest normalized training speed. Hybrid stacks often improve loss, but pure recurrent stacks preserve the strongest long-context iteration-time scaling. The larger DeltaNet and downstream results reinforce the same conclusion: architecture comparisons are meaningful only when scale, optimizer, learning rate, stack pattern, and metric semantics are recorded together.

We also studied a lightweight, linear-time pathway for sharing information across depth in DeltaNet-style stacks. Starting from Cross-Layer Error Residuals, which forward a lower layer's delta-rule write error into the next layer's value target, we found that this natural formulation does not improve over matched DeltaNet or Gated DeltaNet baselines. Diagnosing the cause as a space mismatch, we instead route the signal into the aligned hidden stream through a zero-initialized projection and find that the layer's write \emph{value}, not its write error, is the useful signal to forward. The resulting method, Cross-Layer Value Routing (CLVR), gives a small reduction in final validation loss in the matched single runs reported for both DeltaNet and Gated DeltaNet at 350M parameters, and in the larger Gated DeltaNet run, while preserving the linear-time structure of the host. The gain diminishes as the base model is trained longer or made larger, which leaves its behavior at much larger scale, and on other linear-memory hosts such as Kimi Delta Attention and Gated DeltaNet-2, as the main open question.

\section{Acknowledgments}
Thanks to the Projects in Machine Learning Research (PMLR) course at ETH Zurich, and to CSCS for compute on Alps.

\clearpage
\appendix

\section*{Appendix}
\addcontentsline{toc}{section}{Appendix}

\section{Supplementary 1B-Token Baselines}
\label{app:repo-results}

The main text emphasizes the 15B-token architecture sweep and the larger 1.3B- and 3B-parameter DeltaNet runs because they provide the clearest validation-loss/throughput comparison. Table~\ref{tab:1b-baseline-results} reports an earlier 350M-parameter baseline sweep trained for approximately 1B tokens under the same FineWeb-Edu/LLaMA2 setup, sequence length 4096, and one-node GH200 configuration as the controlled routing experiments.

\begin{table}[tbp]
    \centering
    \caption{Supplementary 350M baseline runs trained for approximately 1B tokens. These shorter-budget runs provide context before the 15B-token sweep but are not used to determine the main validation-loss/throughput frontier.}
    \label{tab:1b-baseline-results}
    \resizebox{\linewidth}{!}{\begin{tabular}{llrrrrr}
\toprule
Baseline & Optimizer & Train loss & Val loss & Val PPL & ktok/s/GPU & TFLOP/s/GPU \\
\midrule
Softmax & AdamW & 2.9111 & 2.9019 & 18.2095 & 144.8 & 334.6 \\
Softmax & Muon & 2.7134 & 2.7240 & 15.2410 & 136.7 & 316.0 \\
Gated DeltaNet & AdamW & 2.8587 & 2.8562 & 17.3950 & 159.5 & 298.6 \\
Gated DeltaNet & Muon & 2.6965 & \textbf{2.7097} & \textbf{15.0254} & 154.8 & 289.6 \\
DeltaNet & AdamW & 2.9484 & 2.9391 & 18.8987 & 153.4 & 296.8 \\
DeltaNet & Muon & 2.6990 & 2.7137 & 15.0847 & 146.7 & 283.7 \\
\bottomrule
\end{tabular}
}
\end{table}

\section{Result Inclusion Criteria}
\label{app:evidence-ledger}

This appendix summarizes how results are separated between the main empirical analysis, supplementary context, and exploratory discussion. Main quantitative claims must identify the model family, stack pattern, scale, token budget, optimizer, learning rate, metric definition, and checkpoint semantics.

\paragraph{Statistical scope.}
The inclusion standard is an auditability standard, not a variance estimate. The reported entries are single runs, so no standard deviations are available; small validation-loss gaps are treated as suggestive unless the same direction appears across matched settings or downstream checks.

\paragraph{Main empirical results.}
The main tables and figures include the 350M-parameter, 15B-token architecture sweep, the learning-rate ablation, sequence-length timing, larger DeltaNet runs, downstream evaluations, the matched 350M-parameter CLER ablations, and the cross-layer routing comparisons (Table~\ref{tab:clvr-results}). These results specify the relevant training scale and metric semantics and are comparable within their experimental groups.

\paragraph{Supplementary context.}
The 350M-parameter, 1B-token baselines in Table~\ref{tab:1b-baseline-results} use a shorter training budget than the main architecture sweep. They provide early baseline and optimizer context but do not determine the final 350M-parameter validation-loss/throughput frontier.

\paragraph{Cross-layer routing results.}
The matched hidden-stream routing comparisons, CLER-H and CLVR, together with the Attention Residuals baseline, are reported in the main text (Table~\ref{tab:clvr-results}) with matched baselines and final validation losses, and meet the same standard as the other main results. The remaining controls, namely surprise-gated value routing, error-plus-value concatenation, output injection, and low-rank projection variants, are reported only as indicative checks; they would need seed-averaged matched runs to enter the main comparison.

\end{document}